\documentclass[letterpaper]{article} 
\usepackage{aaai25}  
\usepackage{times}  
\usepackage{helvet}  
\usepackage{courier}  
\usepackage[hyphens]{url}  
\usepackage{graphicx} 
\usepackage{natbib}  
\usepackage{caption} 
\usepackage{algorithm}
\usepackage{algorithmic}

\usepackage{multirow} 
\usepackage{makecell} 
\usepackage{xcolor}
\usepackage{amsmath}
\usepackage{booktabs}
\usepackage{indentfirst}
\usepackage{soul}
\usepackage{booktabs}
\usepackage{amsmath,amsthm,amssymb,amsfonts}
\usepackage{url}
\usepackage{pifont} 

\usepackage{newfloat}
\usepackage{listings}
\DeclareCaptionStyle{ruled}{labelfont=normalfont,labelsep=colon,strut=off} 
\floatstyle{ruled}
\newfloat{listing}{tb}{lst}{}
\floatname{listing}{Listing}
\title{Harmonious Music-driven Group Choreography with \\ Trajectory-Controllable Diffusion}

\author{
    Yuqin Dai\textsuperscript{\rm 1}, 
    Wanlu Zhu\textsuperscript{\rm 1}, Ronghui Li\textsuperscript{\rm 2}, 
    Zeping Ren\textsuperscript{\rm 2}, Xiangzheng Zhou\textsuperscript{\rm 1}, 
    Jixuan Ying\textsuperscript{\rm 2}, \\ 
    Jun Li\textsuperscript{\rm 1}\thanks{Corresponding authors}, Jian Yang\textsuperscript{\rm 1}\footnotemark[1]
}

\affiliations{
    \textsuperscript{\rm 1}PCA Lab, Key Lab of Intelligent Perception and Systems for High-Dimensional Information of Ministry of Education, School of Computer Science and Engineering, Nanjing University of Science and Technology, Nanjing, China \\
    \textsuperscript{\rm 2}Shenzhen International Graduate School, Tsinghua University, Shenzhen, China \\
    \{daiy, wanluzhu, xzhou, junli, csjyang\}@njust.edu.cn 
    \{lrh22, rzp22, yingjx23\}@mails.tsinghua.edu.cn
    }

\usepackage{bibentry}

\begin{document}

\maketitle

\begin{abstract}
Creating group choreography from music is crucial in cultural entertainment and virtual reality, with a focus on generating harmonious movements. 
Despite growing interest, recent approaches often struggle with two major challenges: \textit{\textbf{multi-dancer collisions}} and \textit{\textbf{single-dancer foot sliding}}.
To address these challenges, we propose a Trajectory-Controllable Diffusion (TCDiff) framework, which leverages non-overlapping trajectories to ensure coherent and aesthetically pleasing dance movements. To mitigate collisions, we introduce a Dance-Trajectory Navigator that generates collision-free trajectories for multiple dancers, utilizing a distance-consistency loss to maintain optimal spacing. Furthermore, to reduce foot sliding, we present a footwork adaptor that adjusts trajectory displacement between frames, supported by a relative forward-kinematic loss to further reinforce the correlation between movements and trajectories.
Experiments demonstrate our method's superiority.
\end{abstract}

%
\begin{links}
    \link{Project Page}{https://wanluzhu.github.io/TCDiffusion/}
\end{links}

\section{Introduction}
\label{sec:intro} 
\noindent Dance, one of the most expressive art forms, has a profound impact on cultural, cinematic, and academic domains \cite{artemyeva2018role, metaverse, dany, xue2024human}.
The process of choreography creation has traditionally been labor-intensive, spurring the development of automated learning models for dance generation. Consequently, music-driven choreography, initially focused on solo dancers \cite{aist++, edge, finedance}, has garnered considerable attention. As the demand for more immersive and interactive experiences grows, the need for multi-person choreography has become increasingly prominent \cite{dany}, leading to a greater emphasis on music-driven group choreography that prioritizes both cohesion and diversity in group movements \cite{schwartz1998passacaille}. However, despite initial recognition and exploration \cite{dany, aioz, gcd, duolando, codancers}, these approaches continue to face two significant challenges:

\begin{figure}[!t]
  \centering
  \includegraphics[width=0.98\linewidth]{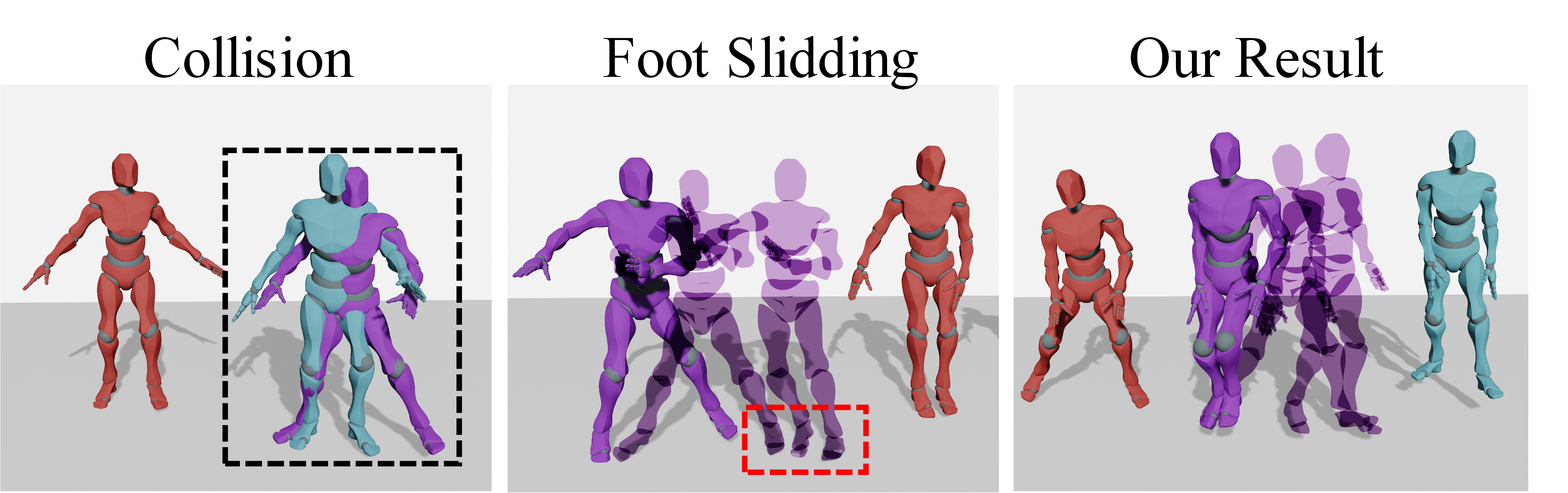}
  \caption{
  Visualizations of two key issues in baseline models: multi-dancer collisions \cite{gcd} (in the \underline{black} box) and single-dancer foot sliding \cite{codancers} (in the \underline{red} box). In contrast, our approach eliminates these issues, delivering superior visual aesthetics.
  }
  \label{fig:intro_challenge}
\end{figure}

\textbf{Multi-dancer collision}.
In many group choreography frameworks \cite{aioz,gcd,dany,edge}, model inputs are typically constructed by concatenating the movements and coordinates of each dancer. However, this strategy results in a significant imbalance, as movements often encompass over 100 dimensions, whereas coordinates are constrained to just three. In group choreography, dancers' coordinates can vary substantially, yet their movements frequently display notable similarities. For example, in the AIOZ-GDance dataset \cite{aioz}, over 80\% of the movements are similar. This similarity leads to \textit{dancer ambiguity}, complicating the model's ability to distinguish between individual dancers, and often leading to collisions, as illustrated in Figure \ref{fig:intro_challenge}.

\textbf{Single-dancer foot sliding}. 
Foot sliding occurs when a dancer’s feet appear to unnaturally glide or shift across the ground, despite accurate movement of the rest of the body, as depicted in the middle of Figure \ref{fig:intro_challenge}. This issue often arises from difficulties in accurately modeling the correlation between the global trajectory and the local rotations of other body parts \cite{longdancediff}. In multi-person choreography, the dancer ambiguity issue further complicates the modeling of this correlation, making it even harder to align footwork with displacement.

To address these challenges, we propose a two-stage method, \textbf{Trajectory-Controllable Diffusion} (TCDiff), which first predicts dancers' coordinates and subsequently generates their movements accordingly.
\textit{\textbf{To mitigate multi-dancer collisions,}} we introduce the Dance-Trajectory Navigator (DTN), designed to resolve dancer ambiguity from representation imbalance by focusing on critical positional coordinates.
This approach centers on a distance-consistency loss that regulates spatial distances, effectively preventing collisions.
Furthermore, we present a simple yet effective fusion projection plug-in that significantly reduces dancer ambiguity while requiring minimal memory.
\textit{\textbf{For single-dancer foot sliding}}, 
We introduce a footwork adaptor that derives foot movements by analyzing displacement between consecutive trajectory frames.
In addition, we propose a Relative Forward-Kinematic (RFK) loss to strengthen the root-motion relationship by enhancing the connection between the root node and the other joints of a dancer.
In summary, our main contributions are:
\begin{itemize}
\item We propose a Dance-Trajectory Navigator that can generate distinct dancer trajectories by exploring a distance-consistency loss to avoid dancer collision. 
\item We introduce a Footwork Adaptor that utilizes trajectory shifts between adjacent frames to generate precise footwork. It incorporates a relative forward-kinematic loss to strengthen the correlation between root node and dance motion, effectively reducing single-dancer foot sliding.
\item Leveraging these components, we develop a novel two-stage multi-dancer generation framework, Trajectory-Controllable Diffusion, to produce high-quality dance movements. Experimental results demonstrate the superiority of our approach over existing methods.
\end{itemize}

\section{Related Work}
\subsection{Music-driven Single-dancer Generation}
\noindent The single-dancer generation is the most relevant area to group dance generation, yet it remains a significant challenge \cite{dancereview}. Early approaches based on motion retrieval paradigms \cite{kovar2002pighin, fan2011example, ofli2011learn2dance, lee2013music} often result in deformed actions. Recent advancements leverage large datasets \cite{lee2019dancing, dancerevolution, valle2021transflower, aist++, finedance, enchantdance} to synthesize motions using deep learning techniques, including auto-regressive models \cite{alemi2017groovenet, yalta2019weakly, ahn2020generative, bailando, xu2024mambatalk} and generative models \cite{mnet, edge, lodge, li2024lodge++}. In recent years, diffusion-based models \cite{diffusion, sohl2015deep, edge, ren2025realistic} have emerged, achieving state-of-the-art performance with high diversity and fidelity.
Solo dance generation prioritizes realism, making artifacts like foot sliding unacceptable \cite{longdancediff}. Existing methods address this by imposing physical constraints via constraint losses and foot contact labels \cite{zhang2021learning, zhang2023real, edge, lodge}.
In addition to generating natural dance movements, enhancing dance action controllability remains an important yet underexplored area. Current methods offer temporal and spatial controls, such as genre \cite{mnet}, text \cite{tm2d,dancecontrol}, and joint control \cite{edge, sinmdm}. 
However, achieving consistent and plausible group dynamics with single-dancer models is challenging, as multi-person dance requires modeling inter-dancer correlations. Additionally, single-dancer models often face issues with dancer ambiguity.

\subsection{Music-driven Multi-dancer Generation}
\noindent  Multi-dancer generation is an emerging area currently in its nascent stage.
To the best of our knowledge, only a handful of studies \cite{dany, aioz, gcd, codancers} have focused on generating scenarios for more than two dancers. 
Among these, GDanceR \cite{aioz} and GCD \cite{gcd} utilize no special structures to avoid the imbalance issue in motion representation, resulting in a tendency for dancer ambiguity. 
CoDancers \cite{codancers} splits group motions into single-person motions, preventing the occurrence of dancer ambiguity. 
However, completely isolating individual features results in incomplete group information, which leads to disharmony generation results. Therefore, a more optimal feature separation strategy is necessary.
In this work, we introduce TCDiff, a method that first generates dancers' trajectories and then produces logical movements. Together with our proposed effective plugin, Fusion Projection, this approach significantly reduces dancer ambiguity.

\subsection{Multi-agent Trajectory Prediction}
\noindent  
To accurately model dancer trajectories, we leverage insights from the trajectory prediction field, which closely aligns with our task by focusing on understanding agent movement.
Past methods heavily rely on hand-crafted rules for describing motions and interactions, including the Gaussian Process~\cite{traj1}, and Markov Models~\cite{traj2}. However, these methods struggle with complex real-world scenarios. Recent deep learning approaches for temporal modeling are LSTM~\cite{traj3} and their variants. To model complex interactions attention-based methods and graph-based approaches have been developed, such as SGCN~\cite{shi2021sgcn} and STAR~\cite{star}. Predicting trajectories is challenging due to the multi-modality issue \cite{mid, socialgan, sociallstm}, where the same input can lead to different outcomes.
Consequently, current methods aim to learn a distribution rather than a single trajectory, employing generative models like GANs~\cite{socialgan} and CVAEs~\cite{ivanovic2020multimodalcvae} and DDPMs~\cite{gu2022stochasticMID}. 
However, dancer trajectories are complicated by reliance on music, adding a new factor and even stationary dancers show intricate positional variations.
In this work, TCDiff utilizes our proposed Dance-Trajectory Navigator, an auto-regressive model that can generate smooth, continuous, and non-overlapping trajectories for dancers.

\begin{figure*}[!t]
  \centering
  \includegraphics[width=0.97\linewidth]{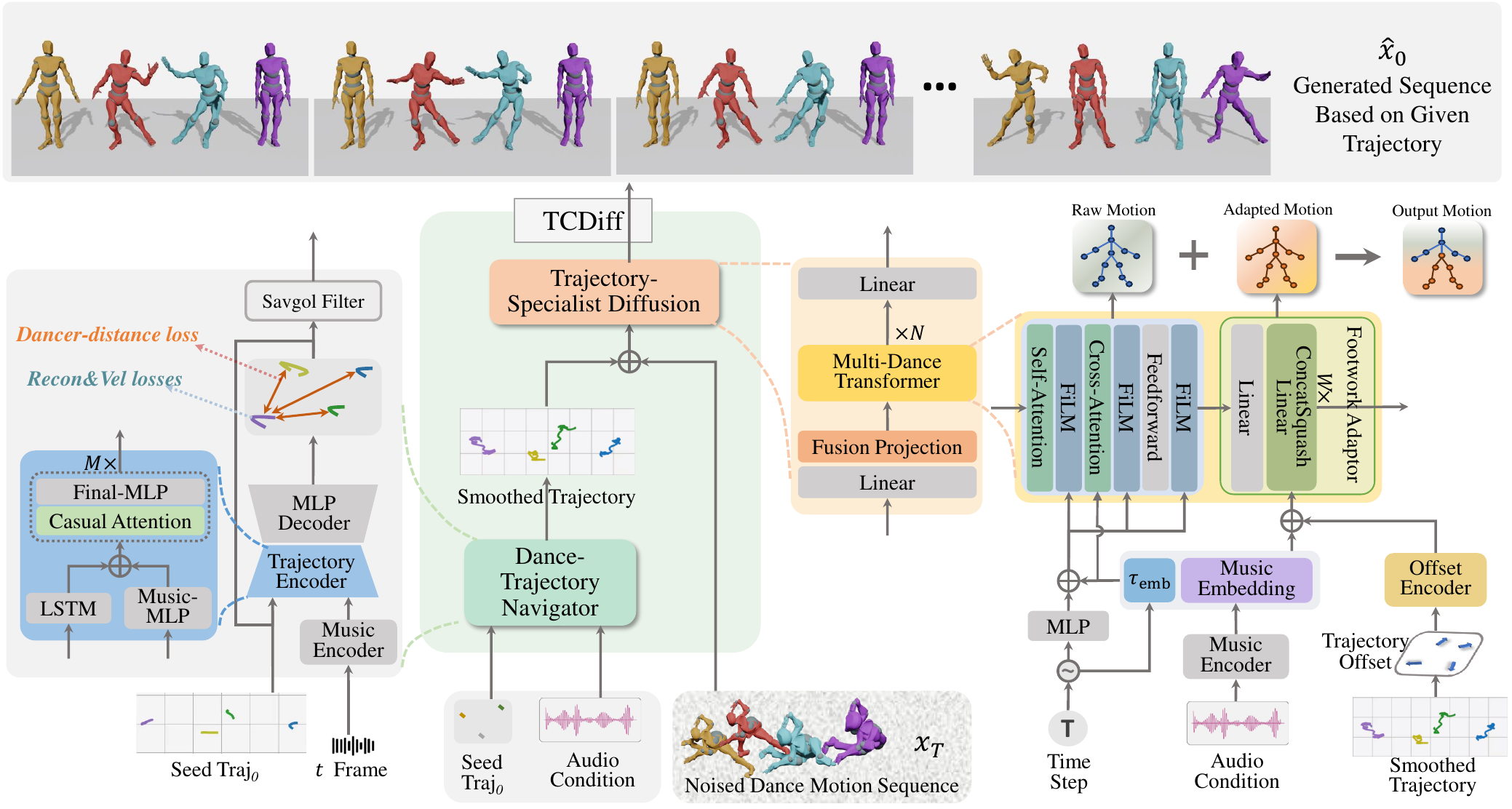}
  \caption{Our TCDiff framework consists of two components: Dance-Trajectory Navigator (DTN) and Trajectory-Specialist Diffusion (TSDiff). Initially, DTN is designed to extract disjoint trajectories (dancer positions) for mitigating dancer ambiguity, as dancers' coordinates exhibit distinct differences and are less prone to confusion.
  Subsequently, TSDiff utilizes the trajectories for conditional diffusion to generate corresponding dance movements. During this process, a Fusion Projection enhances group information before inputting it into the multi-dance transformer, while a footwork adaptor adjusts the final footwork.
  } 
  \label{fig:tcd_framwork}
\end{figure*}

\section{Background}
\label{sec:Background}
\subsection{Music-driven Group Choreography}
\label{sec:ProblemDefinition}
\noindent Given a music sequence $\mathcal{M}=\left\{\small\boldsymbol{m}_i\right\}_{i=1}^L$, group choreography is to generate a corresponding group dance movement sequence $\boldsymbol{x}=\left\{\small\boldsymbol{x}^i\right\}_{i=1}^L$, where  $\boldsymbol{x}^i=\left\{\small\boldsymbol{x}^{i,c}\right\}_{c=1}^C$, $L$, $C$ is the music clip length and the number of dancers, respectively. For simplicity, we use $\boldsymbol{x}_t$ to represent $\{\small\boldsymbol{x}_t^{i,c}\}_{i=1,c=1}^{L,C}$ at $t$ step.

\subsection{Motion and Music Representations} 
\noindent We represent motion features of a dancer utilizing a SMPL~\cite{smpl} pose $\boldsymbol{d} \in \mathbb{R}^{144}$ extracted from a 24-joint SMPL model in 6D rotation format \cite{rot6d}, along with 4-dimensional foot contact labels $\boldsymbol{f} \in \mathbb{R}^{4}$ and a 3-dimensional root node $\boldsymbol{p} \in \mathbb{R}^{3}$ for the positions of the dancer. This results in a motion representation $\boldsymbol{x}=[\boldsymbol{f},\boldsymbol{p},\boldsymbol{d}] \in \mathbb{R}^{151}$. Note that compared to 3D keypoint representations \cite{zhang2021we, zanfir2021thundr, ma20233d}, the use of rotation format tends to yield better motion consistency, as observed in \cite{bailando}.
For the music feature, we follow prior works \cite{mnet, lodge} to utilize Librosa \cite{librosa} to extract a representation $\mathcal{M} \in \mathbb{R}^{35}$, comprising a 1-dimensional envelope, 20-dimensional MFCC, 12-dimensional chroma, along with 1-dimensional one-hot peaks and 1-dimensional one-hot beats.

\subsection{Diffusion Model}
\noindent We generate dance movements via a diffusion-based method \cite{diffusion}, which establishes a Markov noising process that gradually contaminates clean data $\boldsymbol{x}_0$ into standard Gaussian noise $\boldsymbol{x}_T \sim \mathcal{N}(0, \boldsymbol{I})$ through $T$ pollution steps. The corruption process is defined as:
\begin{equation}
q\left(\boldsymbol{x}_t \mid \boldsymbol{x}_{t-1}\right)=\mathcal{N}\left(\sqrt{\alpha_t} \boldsymbol{x}_{t-1},\left(1-\alpha_t\right) \boldsymbol{I}\right),
\end{equation}
where $\alpha_t \in (0,1)$ are pre-defined hyper-parameters, and $\boldsymbol{I}$ is the identity matrix. 
Since the music sequence $\mathcal{M}$ is frequently integrated as a conditioning factor \cite{edge}, the motion creation involves reversing the forward diffusion process by estimating $\widehat{\boldsymbol{x}}_\theta\left(\boldsymbol{x}_t, t, \mathcal{M}\right) \approx \boldsymbol{x}$ with parameters $\theta$ for all $t$. Thus, the basic objective function \cite{diffusion} is defined as:
\begin{equation}
\label{eq:simpleloss}
\mathcal{L}_{\text {simple }}=\mathbb{E}_{\boldsymbol{x}, t}\left[\left\|\boldsymbol{x}-\widehat{\boldsymbol{x}}_\theta\left(\boldsymbol{x}_t, t, \mathcal{M}\right)\right\|_2^2\right].
\end{equation}
We extend the diffusion model by incorporating dancers' trajectories to produce more realistic dance movements.

\section{Methodology}
\label{sec:method}
\noindent In this section, we introduce our \textbf{T}rajectory-\textbf{C}ontrollable \textbf{Diff}usion (\textbf{TCDiff}) framework, which can generate synchronized group dance movements from a music clip. 
Our pipeline consists of a Dance-Trajectory Navigator \textbf{(DTN)} and a Trajectory-Specialist Diffusion \textbf{(TSDiff)} in Figure \ref{fig:tcd_framwork}. 

\subsection{Dance-Trajectory Navigator}
\label{sec:DancebeatNavigator}
\noindent To mitigate dancer ambiguity resulting from similar movements among dancers, our Dance-Trajectory Navigator (DTN) module aims to suppress the interference of similar movements and prioritize the modeling of coordinates, as shown in the left of Figure~\ref{fig:tcd_framwork}. Starting with the music sequence $\mathcal{M}=\left\{\boldsymbol{m}_i\right\}_{i=1}^L$ and a seed trajectory $\boldsymbol{p}^0=\{\boldsymbol{p}^{0,1},\ldots,\boldsymbol{p}^{0,C}\}$, the DTN module takes the coordinates $\boldsymbol{p}^i=\{\boldsymbol{p}^{i,1},\ldots,\boldsymbol{p}^{i,C}\}$ and the music frame $\boldsymbol{m}_i$ at sequence step $i$ as inputs to a music encoder \cite{mnet} and a trajectory encoder, along with an MLP decoder, to recursively generate the next dancer coordinates $\widehat{\boldsymbol{p}}^{i+1}=\{\widehat{\boldsymbol{p}}^{(i+1),1},\ldots,\widehat{\boldsymbol{p}}^{(i+1),C}\}$. The final trajectory $\widehat{\boldsymbol{p}}=\{\widehat{\boldsymbol{p}}^1,\ldots,\widehat{\boldsymbol{p}}^L\}$ is obtained by applying a Savgol filter \cite{savgol} to smooth the dancer coordinates. The MLP decoder is implemented as a simple multilayer perceptron. Next, we describe the trajectory encoder. 

\textbf{Trajectory Encoder} is tasked with extracting detailed features from both the input music and coordinate sequences, which are then fed into the MLP decoder to generate trajectory predictions. For feature pre-processing, the music sequence undergoes processing by a music-MLP, while the coordinate sequence is inputted into a sequence model \cite{lstm} to extract temporal features. Additionally, instead of relying on absolute positional encoding \cite{abpos1, attention} or naive positional encoding \cite{repos1, repos2}, we utilize identity encoding (IE) and temporal positional encoding (TPE) \cite{tbiformer} to capture temporal information.
Furthermore, we introduce a trajectory attention module comprising Casual Attention \cite{attention,causalAtten}, effectively directing the model's focus to past information through masking, in conjunction with an MLP network as:
\begin{equation}
\operatorname{Attn}=\operatorname{Softmax}\left(\boldsymbol{M} \boldsymbol{P}^T / \sqrt{d}+\boldsymbol{B}\right) \boldsymbol{P} \times mask,
\end{equation}
where $\boldsymbol{M}$, $\boldsymbol{P}$ are the processed music and position features using both the music-MLP and the LSTM, respectively. 
The $mask$ is the causal mask with $mask_{i,j} = -\infty \times 1(i > j) + 1 (i \leq j)$, where $1(\cdot)$ is the indicator function.
$\boldsymbol{B}$ is the bias, and $d$ is a scaling factor to ensure the stability of the model's training process. We replicate the trajectory attention module $M$ times in this context.

\textbf{DTN Loss.}
\label{sec:DTNLossTerm}
The objectives of the DTN module are determined by some observations that the ground truth typically exhibits characteristics of continuity and non-overlap.
However, it is insufficient to guarantee this if relying solely on the reconstruction loss $\mathcal{L}_{\text {recon}}$ \cite{srlstm}. Therefore, we apply a velocity loss $\mathcal{L}_{\text {v}}$ \cite{edge} to restrict position variation for ensuring the continuity. To further approximate the ground truth for reducing the overlapping prediction, simultaneously, we introduce a distance-consistency loss $\mathcal{L}_{DC}$,
\begin{align}
\Delta\boldsymbol{p}^{w,ij}= \left(\boldsymbol{p}^{w,i}-\boldsymbol{p}^{w,j}\right)-\left(\widehat{\boldsymbol{p}}^{w,i}-\widehat{\boldsymbol{p}}^{w,j}\right), \\
\mathcal{L}_{DC}=\frac{1}{C-1} \sum_{w=1}^L\binom{C}{2}_{ij}\left\|\Delta\boldsymbol{p}^{w,ij}\right\|_2^2,
\end{align}
which ensures that the spacing among dancers is within an appropriate range.
The overall loss $\mathcal{L}_{dtn}$ for DTN is $\mathcal{L}_{dtn}=\mathcal{L}_{recon}+\lambda_{v}\mathcal{L}_{v}+\lambda_{DC}\mathcal{L}_{DC}$, where $\lambda_{v}$ and $\lambda_{DC}$ are the balanced hyper-parameters. 

\subsection{Trajectory-Specialist Diffusion}
\label{sec:TSDiff}
\noindent After extracting disjoint trajectories from the DTN module, we introduce a Trajectory-Specialist Diffusion (TSDiff) to generate dancer movements characterized by non-overlapping steps and enhanced grip, in accordance with the provided trajectories. TSDiff consists of a Fusion Projection, a multi-dancer transformer, and simple linear layers for input and output. 
To ensure non-overlapping movements, we leverage the disjoint trajectories for conditional denoising.
To mitigate foot-slide, we present a footwork adaptor within the multi-dancer transformer to adjust foot movements based on trajectory information, thereby reducing foot-slide occurrences. To fully exploit the known trajectory feature, we refrain from introducing noise to the provided positional information $\widehat{\boldsymbol{p}}$, opting instead for conditional motion denoising. Since the linear layers are simple single linear networks, the Fusion Projection and Multi-Dancer Transformer modules are as follows:

\begin{figure}[!t]
  \centering
  \includegraphics[width=0.97\linewidth]{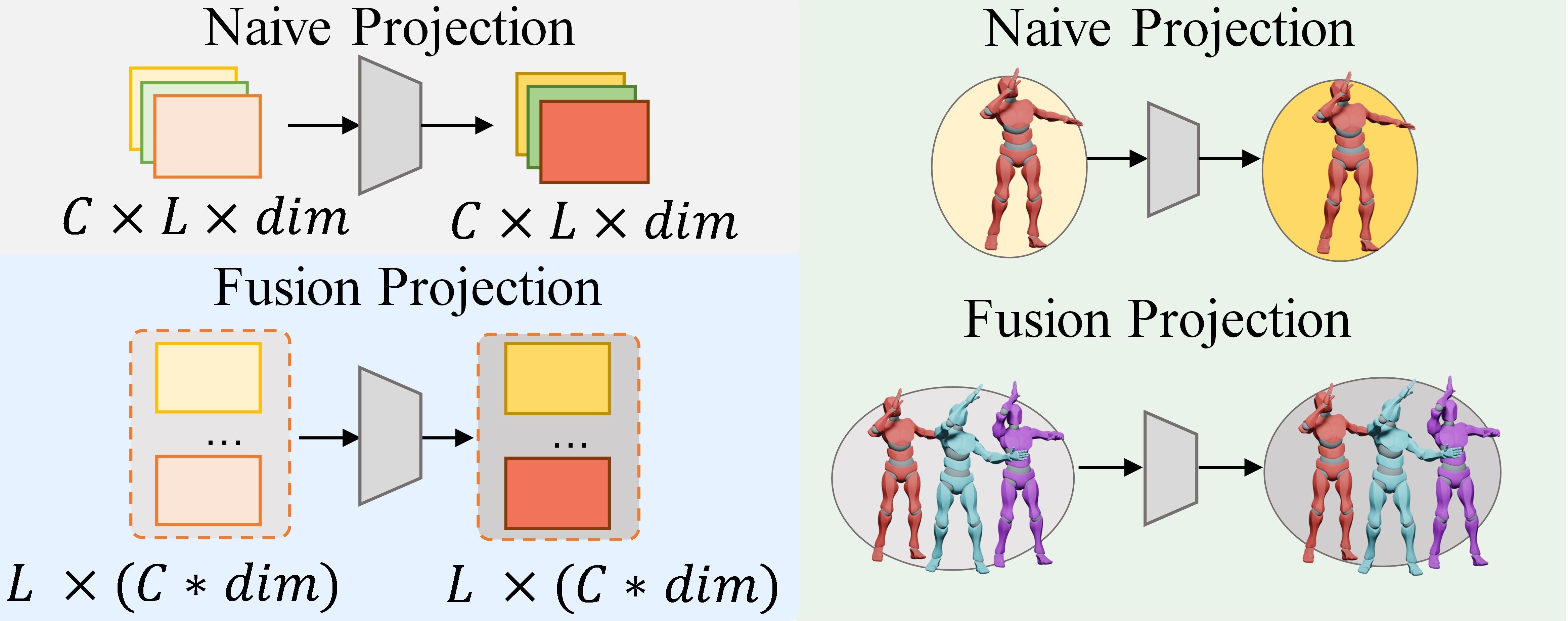}
  \caption{Fusion Projection Module.
  }
  \label{fig:fp}
\end{figure}

\textbf{Fusion Projection (FP). }
\label{sec:FusionProjection}
We propose a simple yet effective solution that tackles dancer ambiguity challenges via its unique feature processing, as shown in Figure~\ref{fig:fp}. 
The primary rationale behind the FP module is that high-dimensional inputs can capture more distinctive features. By increasing the dimensionality of input features through stacking, we construct high-dimensional input nodes for an expanded MLP, effectively reducing ambiguity among dancers. 
Relevant evidence is provided in the ablation study.

\textbf{Multi-Dancer Transformer}
\label{sec:Multi-DancerTransformer}
employs an initial decoder block $D(\cdot)$ \cite{edge} to generate raw motion, which is subsequently refined by our Footwork Adaptor to achieve realistic foot placement. 
To be more specific, given frame-level music information $\mathcal{M}=\{\boldsymbol{m}_i\}_i^L$, the diffusion final time step $T$, and the offset of the given trajectory $\mathcal{V}=\{\boldsymbol{v}_i\}_i^L$ (e.g., velocity) as conditions, we first utilize $D(\cdot)$ for feature processing to derive raw motion $\boldsymbol{\widehat{r}}$: 
\begin{equation}
    \boldsymbol{\widehat{r}}=D(\boldsymbol{x}_T,T,\mathcal{M}).
\end{equation}
Subsequently, we propose \textit{\textbf{Footwork Adaptor}} module $FA(\cdot)$ for correction to obtain the adapted motion $\boldsymbol{\widehat{a}}$.
\begin{equation}
\boldsymbol{\widehat{a}}=FA(\boldsymbol{\widehat{r}},T,\mathcal{M}, \mathcal{V}).
\end{equation}
Given that positional changes are predominantly driven by footwork, our approach focuses exclusively on modulating the dancers' lower body movements.
Therefore, we split $\boldsymbol{\widehat{r}}$ and $\boldsymbol{\widehat{a}}$ into upper and lower body $\boldsymbol{\widehat{r}}=\{\boldsymbol{\widehat{r}}_{upper},\boldsymbol{\widehat{r}}_{lower}\}$, and $\boldsymbol{\widehat{a}}=\{\boldsymbol{\widehat{a}}_{upper},\boldsymbol{\widehat{a}}_{lower}\}$. The down part that is closely related to footwork is picked as the final generated result:
\begin{equation}
\widehat{x}_0=\boldsymbol{\widehat{r}}_{upper} \oplus \boldsymbol{\widehat{a}}_{lower}.
\end{equation}
The Footwork Adaptor consists of a linear layer and a ConcatSquashLinear layer, as shown in Figure \ref{fig:tcd_framwork}, which has been proven to be effective in various coordinate prediction domains \cite{mid, 3dpointdiffusion}.

\textbf{Conditional Motion Denoising.}
To leverage the provided trajectory information, instead of adding noise to all information like the original diffusion does, we exclude adding noise to trajectory information. 
For denoising, we do not denoise $\widehat{\boldsymbol{p}}$ in $\boldsymbol{x}=[\boldsymbol{d},\boldsymbol{f},\widehat{\boldsymbol{p}}]$, but treat it as a condition for motion denoising.
At each step $t$ in forward processing, we concatenate the noised dance pose information $\boldsymbol{d{}_t}$, the noised contact label $\boldsymbol{f}_t$ and the given position $\mathcal{\widehat{\boldsymbol{p}}}$ into one vector:
\begin{equation}
\boldsymbol{x}_t= \mathcal{\boldsymbol{d{}_t} \oplus \boldsymbol{f{}_t} \oplus \widehat{\boldsymbol{p}}}.
\end{equation}
Because the above data format is consistent with the original, we can still use simple loss in Eq. \ref{eq:simpleloss} for optimization.
At this point, motion $\boldsymbol{d}$ and contact label $\boldsymbol{f}$ are denoised while coordinates $\widehat{\boldsymbol{p}}$ are reconstructed. This enhances the model's learning and memory capabilities for coordinates while improving its ability to extract features from trajectories.

\textbf{TSDiff Loss.}
\label{sec:losses}
To enable trajectory-conditional generation, we introduce the Relative Forward-Kinematic (RFK) loss $\mathcal{L}_{\mathrm{RFK}}$ to enhance root-motion correlation. The $\mathcal{L}_{\mathrm{RFK}}$ adjusts the individual dancers' relative distance between root nodes and other body joints and can be formulated as:
\begin{equation}
\footnotesize
\mathcal{L}_{\mathrm{RFK}}=\frac{1}{L} \sum_{i=1}^L\left\|\left(\text{FK}\left(\boldsymbol{d}\right) - \text{FK}\left(\boldsymbol{p}\right)\right) 
-(\text{FK}(\widehat{\boldsymbol{d}})-\text{FK}(\widehat{\boldsymbol{p}}))\right\|_2^2.
\end{equation}
Here, $\text{FK}(\cdot)$ is the forward kinematic function that calculates the positions of joints given the 6D rotation motion.
We adopt joint velocity loss $\mathcal{L}_{\text {vel }}$ and the foot contact loss $\mathcal{L}_{\text {contact }}$ from \cite{edge}. 
The overall objective of our proposed TSDiff is
$\mathcal{L}_{\mathrm{TSDiff}}=\mathcal{L}_{\text {simple }}+\lambda_{\text {RFK}} \mathcal{L}_{\text {RFK }}+\lambda_{\text {vel }} \mathcal{L}_{\text {vel }}+\lambda_{\text {contact }} \mathcal{L}_{\text {contact }}$, where $\lambda_{\text {RFK}} $, $\lambda_{\text {vel }}$ and $\lambda_{\text {contact }}$ are the balanced hyper-parameters.

\section{Experiments}
\label{sec:Experiments}
\subsection{Experimental Settings}
\noindent \textbf{Implementation Details.}
\label{sec:ImplementationDetails}
For our Dance-Trajectory Navigator, the $\lambda_{v}=\lambda_{DC}=2$, and the hidden size of all module layers is set to 64. 
The Trajectory transformer, which is stacked with $M = 6$ transformer layers, is equipped with 8 heads of attention. The $\lambda_{\text {RFK}}=0.6$, $\lambda_{vel}=3$, the $\lambda_{joint}=0.6$, and the $\lambda_{contact }=10$. Both the LSTM model and the Music-MLP consist of 3 layers each. The Final-MLP processes the information passed to it through 4 layers, utilizing LeakyReLU non-linearity as the activation function.  The sequence length $L = 120$, the hidden dimension is 512, with $N = 8$ layers and 8 heads of attention. 
We apply a 3-layer MLP as a Fusion Projection, followed by ReLU activation at each layer. Additionally, we stack $W = 3$ Concat Squash Linear with a hidden size of $d_{csl} = 128$ and $d_{ctx} = 512$.
The entire framework was trained on 4 Nvidia 4090 GPUs for 3 days. We use a single 4090 GPU to train the Dance-Trajectory Navigator for 26 hours, utilizing batch sizes of 750, 400, 256, and 170 for 2, 3, 4, and 5 dancers, respectively. Similarly, the TSDiff model was trained on 4 NVIDIA 4090 GPUs for 2 days, employing batch sizes of 60, 53, 32, and 20, in that order.

\textbf{Dataset.}
\label{sec:Dataset}
AIOZ-GDance dataset \cite{aioz} is an extensive repository of group dance performances comprising 16.7 hours of synchronized music and 3D multi-dancer motion data. This dataset encompasses a diverse array of over 4000 dancers, spanning 7 distinct dance styles and 16 music genres. Following the partition setting \cite{aioz}. we randomly sample all videos into train, validation and test sets with 80\%, 10\% and 10\% of total videos, respectively.

\begin{figure}[t]
  \centering
  \includegraphics[width=0.97\linewidth]{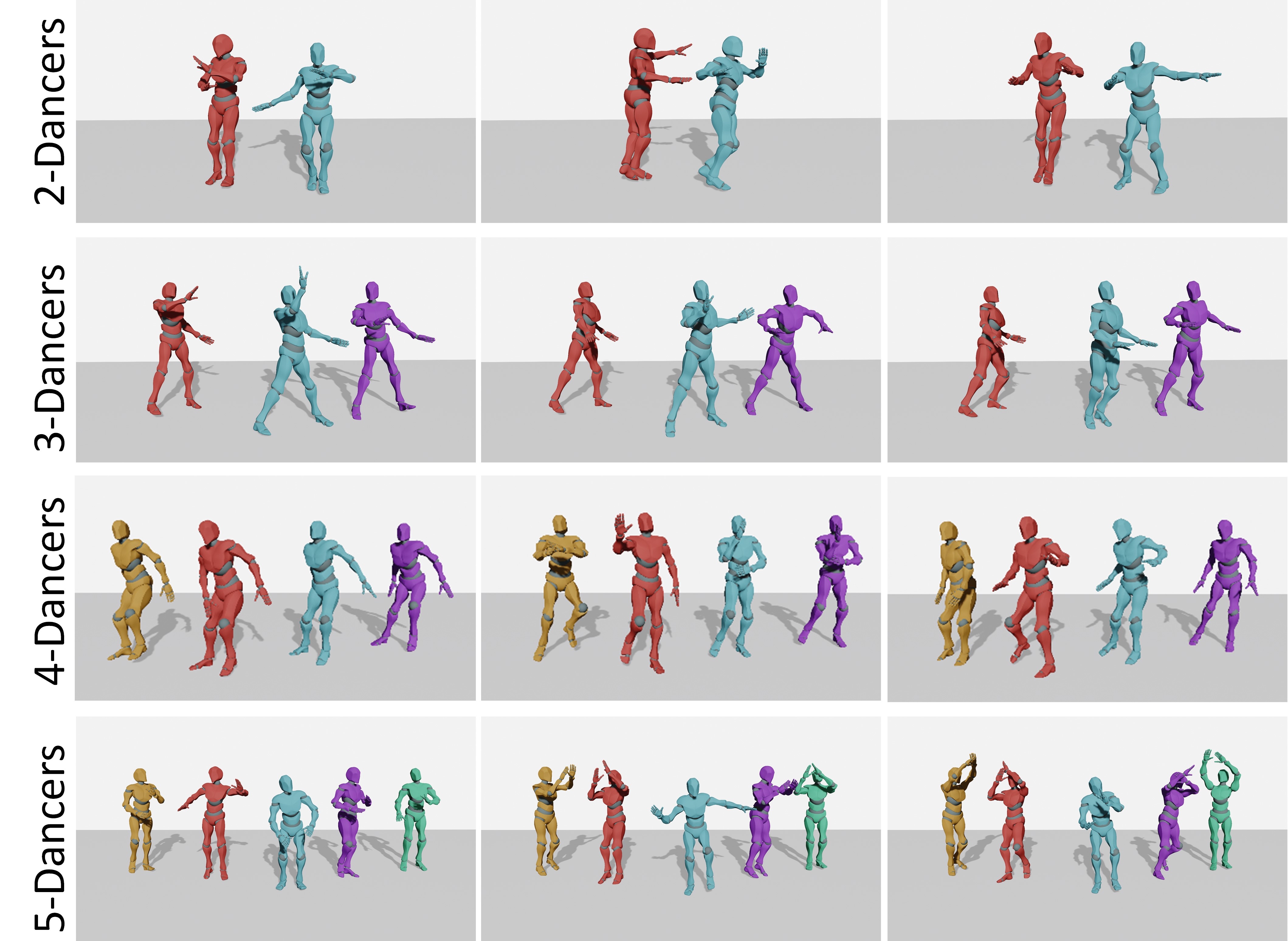}
  \caption{Generated results with different dancer counts. 
  }
  \label{fig:various_dancer}
\end{figure}

\textbf{Compared methods.}
\label{sec:baselines}  
We compare our proposed MotionDiffuse with three baseline models: GDanceR \cite{aioz}, GCD \cite{gcd}, and CoDancers\cite{codancers}.
To the best of our knowledge, these represent all the available group dance generation models capable of producing choreography for two or more dancers.
Additionally, we incorporate EDGE \cite{edge}, the prevailing single-dancer model, to underscore our approach by training it on the AIOZ-GDance dataset.

\begin{figure*}[tb]
  \centering
  \includegraphics[width=0.90\linewidth]{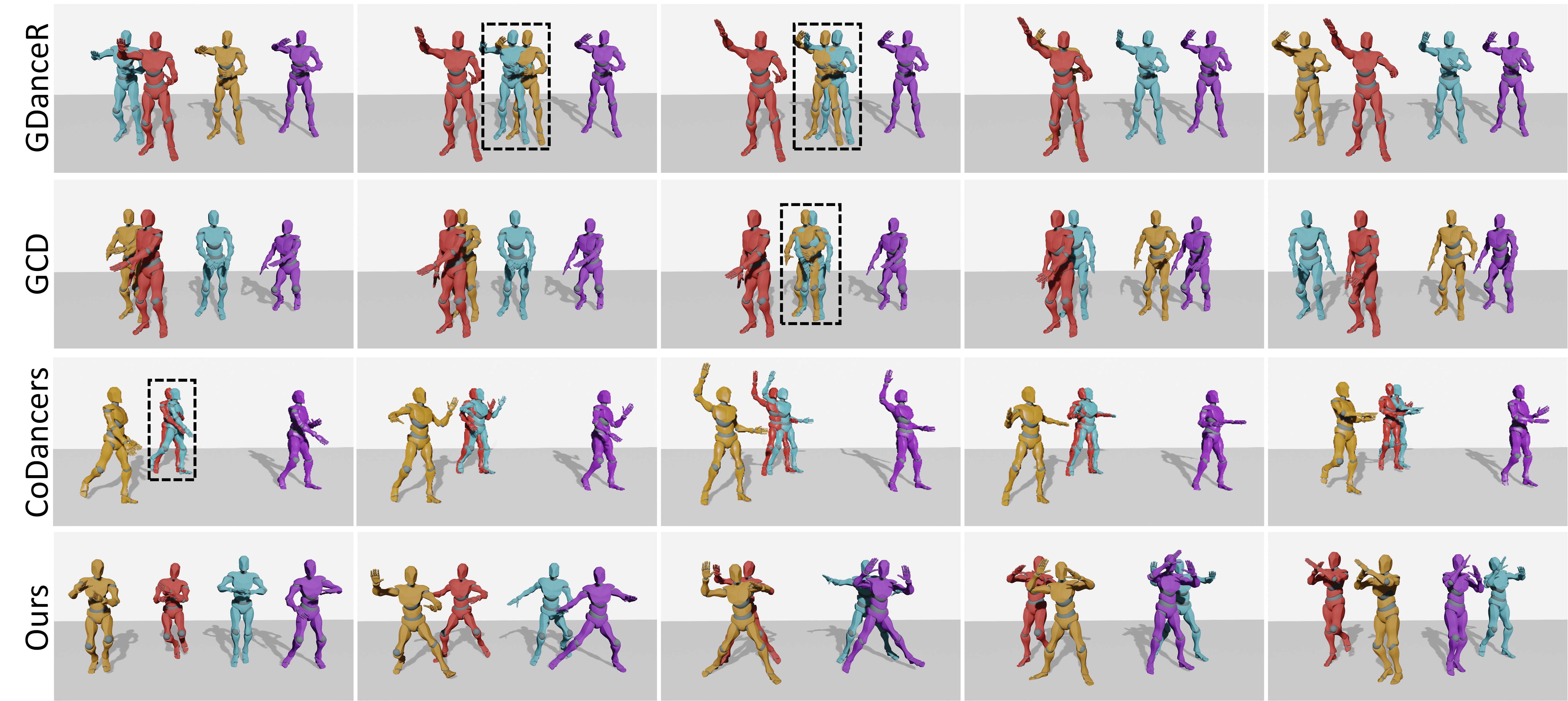}
  \caption{Visual comparison with Baselines. These methods often result in collisions (in \underline{black} box) or lack of foot movements during exchanges. In contrast, our model minimizes dancer overlaps and generates more natural footwork. 
  }
  \label{fig:vis_comparison}
\end{figure*}
\textbf{Metrics.}
\label{sec:Metrics}
We employ metrics for both multi-dancer and single-dancer evaluations to assess our model. For multi-dancer assessment \cite{aioz}. Group Motion Realism (GMR) measures feature similarity via Frechet Inception Distance, Group Motion Correlation (GMC) evaluates coherence through cross-correlation between generated dancers, and Trajectory Intersection Frequency (TIF) assesses the frequency of collisions among dancers.
For single-dancer evaluation, Frechet Inception Distance (FID) \cite{aist++, heusel2017gans} quantifies the similarity between individual dances and ground-truth dances. Generation Diversity (Div) \cite{aist++, dancerevolution} appraises the variety of dance movements using kinetic features. Motion-Music Consistency (MMC) \cite{aist++} evaluates how well generated dances synchronize with the music beat. Physical Foot Contact score (PFC) \cite{edge} indicates the physical plausibility of footwork by considering the correlation between the center of mass and foot velocity.


\subsection{Comparison to the State of the Art}
\label{sec:Comparison}

\begin{figure}[tb]
  \centering
  \includegraphics[width=0.97\linewidth]{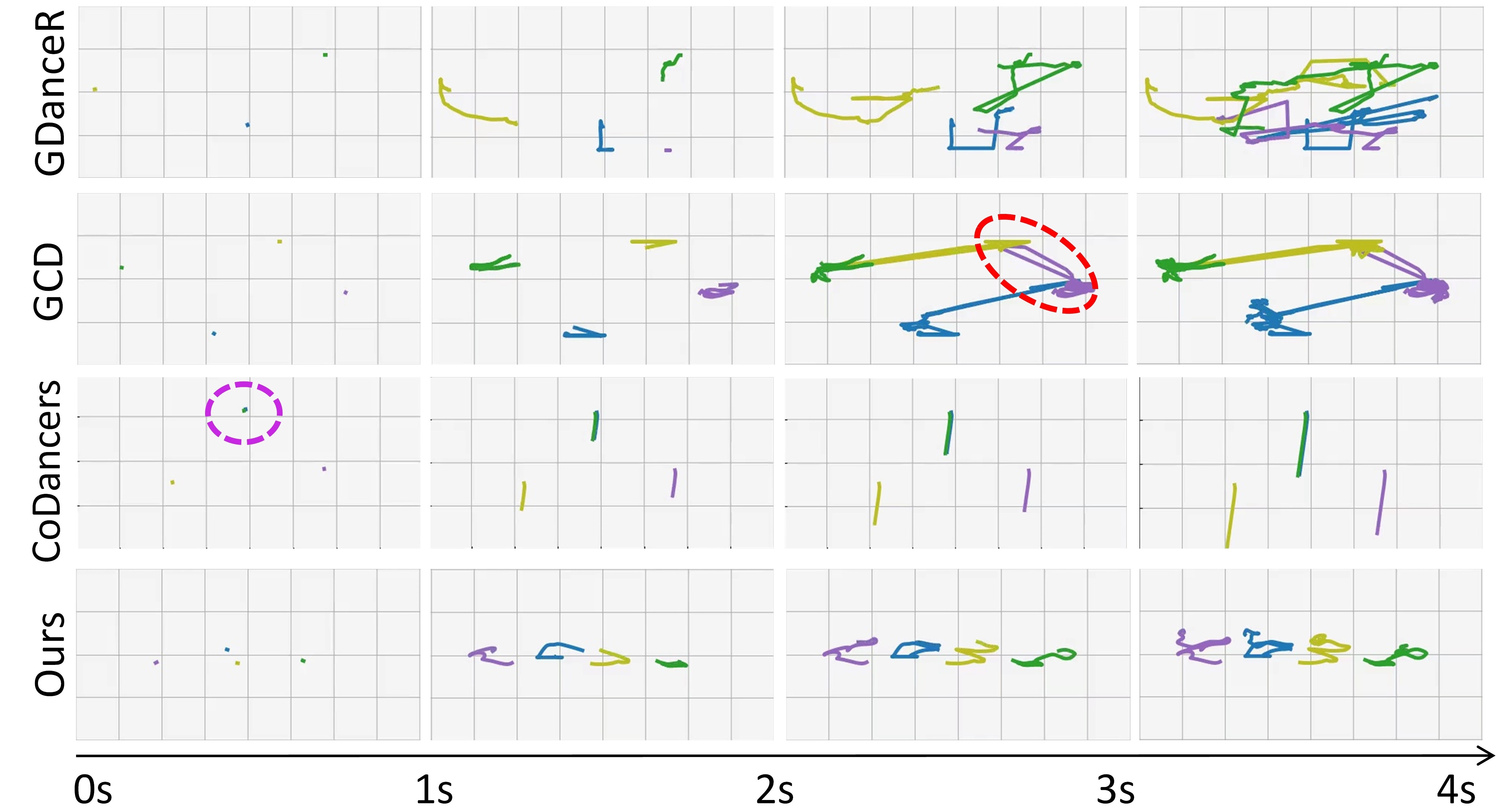}
  \caption{Top-down comparisons of dancer trajectories.
  GDanceR produces overlapping paths, while GCD causes an unnatural shift, returning to the starting point and ultimately overlapping with a blue dancer. CoDancers generates unreasonable initial positions for the dancers, leading to significant overlap. In contrast, our model effectively minimizes overlaps, showcasing the superior performance of our DTN.
  }
  \label{fig:traj}
\end{figure}


\begin{table}[tb]
  \centering
  \setlength{\tabcolsep}{1.2pt} 
  \begin{tabular}{lcccccccc}
    \toprule
    \multirow{2}{*}{Method} & \multicolumn{3}{c}{Group-dance Metric} & \multicolumn{4}{c}{Single-dance Metric} \\
    \cmidrule(lr){2-4} 
    \cmidrule(lr){5-8}
    & GMR$\downarrow$ & GMC$\uparrow$ & TIF$\downarrow$ & FID$\downarrow$ & Div$\uparrow$ & MMC$\uparrow$ & PFC$\downarrow$  \\
    \midrule
    EDGE & 63.35 & 61.72 & 0.36 & 31.40 & 9.57 & 0.26 & 2.63 \\
    GDanceR & 51.27 & 79.01 & 0.22 & 43.90 & 9.23 & 0.25 & 3.05 \\
    GCD & 31.47 & \underline{80.97} & 0.17 & \underline{31.16} & \underline{10.87} & \textbf{0.26} & 2.53  \\
    CoDancers & \underline{26.10} & 74.05 & \textbf{0.10} & \textbf{23.98} & 9.48 & 0.25 & 3.26 \\
    \midrule
    \textbf{Ours} & \textbf{13.86} & \textbf{81.98} & \underline{0.13} & 37.47 & \textbf{15.10} & \underline{0.25} & \textbf{0.51} \\
    \bottomrule
  \end{tabular}
  \caption{
  Quantitative comparison with the baselines. 
  }
  \label{tab:baselines}
\end{table}

\noindent \textbf{Qualitative Visual Comparison.}
The performance of our model is illustrated in Figures \ref{fig:various_dancer} and \ref{fig:vis_comparison}, highlighting its ability to generate aesthetically pleasing results across various group sizes. Top-view dancer trajectories in Figure \ref{fig:traj} further demonstrate our model's superiority in minimizing overlaps. 
In contrast, GDanceR ~\cite{aioz} and GCD~\cite{gcd} overly prioritize movement similarity, neglecting positional differences, which leads to dancer ambiguity and confusion in positioning. CoDancers~\cite{codancers} produces unreasonable initial positions due to incomplete group information, resulting in severe overlaps. Additionally, baseline methods suffer from dancer ambiguity, limiting their ability to align footwork actions with positional changes, thereby hindering the generation of accurate footwork.


\begin{figure}[!t]
  \centering
  \includegraphics[width=0.97\linewidth]{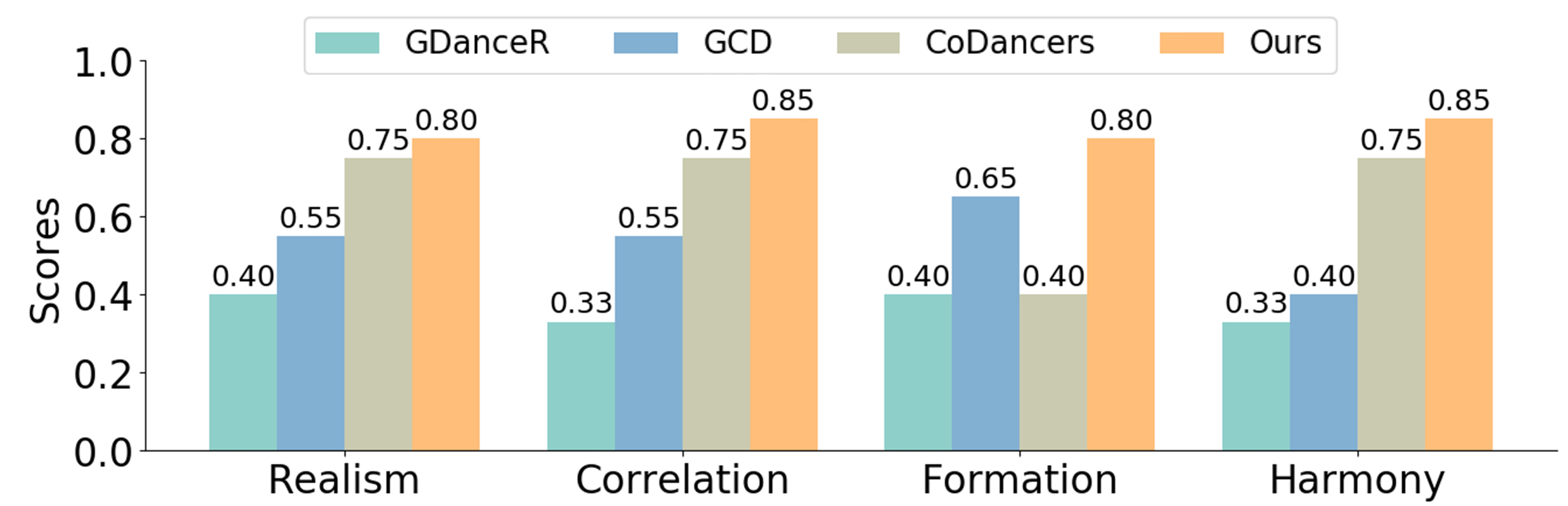}
  \caption{User study based on four criteria: motion realism, music-motion correlation, formation aesthetics, and harmony of dancers. Our model has garnered greater user favor, showcasing our superiority in aesthetic appeal.
  }
  \label{fig:userstudy}
\end{figure}

\textbf{Quantitative Results.}
Tables \ref{tab:baselines} and \ref{tab:dancernummetrics} compare our model's performance with baseline methods. 
Our model consistently outperforms in group-dance metrics and excels in Div and PFC for single-dance Metrics. TCDiff effectively captures inter-dancer correlations (high GMC) with a slight trade-off in individual fidelity (FID). Leveraging enriched group dance information, it generates realistic and harmonious multi-person sequences (low GMR), boosting diversity and quality.
In contrast, single-dancer models like EDGE~\cite{edge} perform well on solo metrics but struggle in multi-person scenarios, showing high TIF due to dancer ambiguity. GDanceR~\cite{aioz} produces low-quality movements (high TIF), reflecting poor ambiguity handling. Similarly, GCD~\cite{gcd} suffers from unnatural transitions, leading to high TIF and limited coherence.
CoDancers~\cite{codancers} reduces ambiguity (low TIF) but compromises inter-dancer correlations (low GMC) and formation integrity, resulting in discordant group formations, as shown in our user study (Figure \ref{fig:userstudy}).
By decoupling dancer coordinates and movements into two stages, our method mitigates ambiguity, improves coordination, and enhances formation quality.

\begin{table}[!t]
  \centering
  \setlength{\tabcolsep}{1.5pt} 
  \begin{footnotesize}
  \begin{tabular}{lccccccccc}
    \toprule
     Method & Dancers\# & GMR$\downarrow$ & GMC$\uparrow$ & TIF$\downarrow$ &  FID$\downarrow$ & Div$\uparrow$ & MMC$\uparrow$ \\
    \midrule
   \multirow{4}{*}{GDanceR} & 2 & 53.83 & 75.44 & 0.286 & 48.82 & 9.36 & 0.248 \\
    & 3 & 55.85 & 74.07 & 0.204 & 44.47 & 9.36 & 0.245 \\
    & 4 & 58.79 & 77.71 & 0.162 & 47.32 & 9.24 & 0.248 \\
    & 5 & 55.05 & 78.72 & 0.218 & 44.19 & 8.99 & 0.249 \\
    \midrule
    \multirow{4}{*}{GCD} & 2 & 34.09 & 80.26 & 0.167 & 32.62 & 10.41 & 0.266 \\
    & 3 & 36.25 & 79.93 & 0.184 & 33.94& 10.02 & 0.266 \\
    & 4 & 36.28 & 81.82 & 0.125 & 35.89 & 9.87 & 0.251 \\
    & 5 & 38.43 & 81.44 & 0.168 & 35.08 & 9.92 & 0.264 \\
    \midrule
    \multirow{4}{*}{CoDancers} & 2 & 24.53 & 72.88 & 0.080 & 26.31 & 9.01 & 0.251 \\
    & 3 & 27.23 & 74.34 & 0.084 & 24.85 & 9.15 & 0.254 \\
    & 4 & 26.44 & 75.34 & 0.097 & 25.76 & 9.43 & 0.258 \\
    & 5 & 26.34 & 74.22 & 0.113 & 25.45 & 9.77 & 0.253 \\
    \midrule
    \multirow{4}{*}{\textbf{Ours}} & 2 & 15.77 & 81.92 & 0.121 & 41.26 & 16.20 & 0.263 \\
    & 3 & 10.97 & 81.51 & 0.123 & 48.00 & 19.28 & 0.253 \\
    & 4 & 13.44 & 81.70 & 0.149 & 23.32 & 10.89 & 0.253 \\
    & 5 & 15.36 & 82.77 & 0.109 & 37.31 & 14.01 & 0.236 \\
    \bottomrule
  \end{tabular}
  \caption{Comprehensive comparison with SOTA methods across varying numbers of dancers.
  } 
  \label{tab:dancernummetrics}
  \end{footnotesize}
\end{table}

\begin{table}[!t]
  \setlength{\tabcolsep}{1.5pt} 
  \centering
  \begin{footnotesize}
  \begin{tabular}{lcccccccc}
    \toprule
    \multirow{2}{*}{Method} & \multicolumn{3}{c}{Group-dance Metric} & \multicolumn{4}{c}{Single-dance Metric} \\
    \cmidrule(lr){2-4} 
    \cmidrule(lr){5-8}
    & GMR$\downarrow$ & GMC$\uparrow$ & TIF$\downarrow$ & FID$\downarrow$ & Div$\uparrow$ & MMC$\uparrow$ & PFC$\downarrow$  \\
    \midrule
    w/o CMD & 33.25 & 80.61 & 0.148 & 43.77 & 15.22 & 0.23 & 0.95  \\
    w/o FA and FP & 25.57 & 74.80 & 0.149 & 27.18 & \textbf{17.86} & 0.19 & 4.29  \\
    w/o FA & 21.39 & 80.13 & 0.149 & 29.21 & 12.79 & 0.22 & 3.25  \\
    w/o FP & 24.25 & \textbf{82.63} & 0.148 & \textbf{23.77} & 16.00 & 0.21 & 0.72  \\
    Full & \textbf{13.86} & 81.98 & \textbf{0.126} & 37.47 & 15.10 & \textbf{0.25} & \textbf{0.51} \\
    \bottomrule
  \end{tabular}
  \caption{Ablation study of Conditional Motion Denoising (CMD), Footwork Adaptor (FA) and Fusion Projection (FP).} 
  \label{tab:ablationstudy}
  \end{footnotesize}
\end{table}

\begin{figure}[!t]
  \centering
  \includegraphics[width=0.95\linewidth]{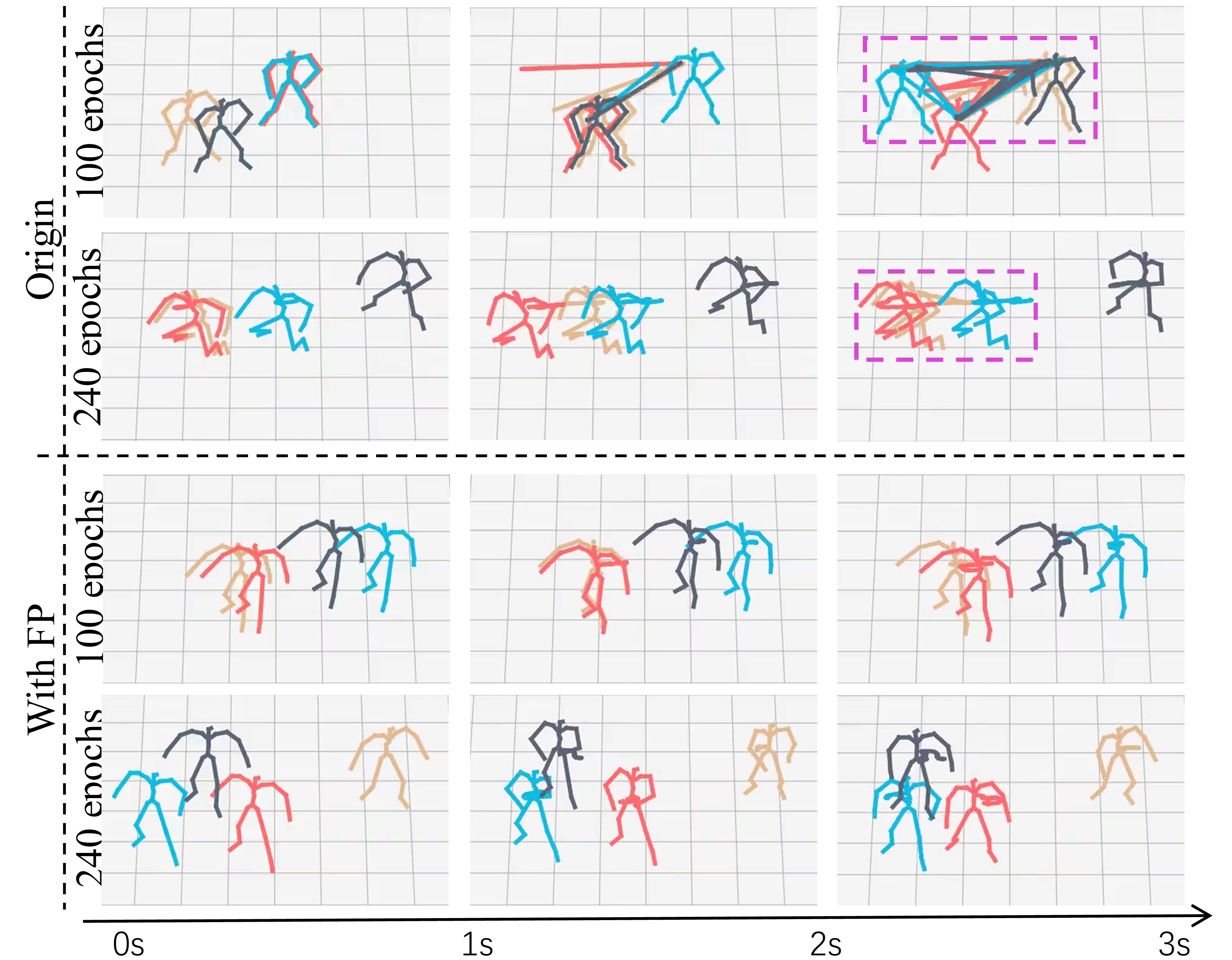}
  \caption{Comparison of EDGE results with and without our FP module. The figure displays various dancers and their movement trajectories. Incorporating the FP module significantly reduces overlap and irregular swapping phenomena.
  }
  \label{fig:fusion_projection}
\end{figure}

\subsection{Ablation Study}
\label{sec:ablationstudy}
\noindent \textbf{Effectiveness of Conditional Motion Denoising.}
Table \ref{tab:ablationstudy} shows that our Conditional Motion Denoising (CMD) improves GMR, GMC, FID, MMC, and PFC metrics. CMD effectively converts trajectory denoising loss into reconstruction loss, enhancing generation quality and maximizing the use of trajectory features. Additionally, clean trajectory data enables more accurate computation of the RFK loss and the FA, further boosting the model's trajectory-based generation performance.

\textbf{Effectiveness of Fusion Projection.}
\label{sec:fpab}
As shown in Table \ref{tab:ablationstudy}, FP noticeably improves metrics such as GMR, GMC, MMC, and PFC, leading to more lifelike group dance sequences with better synchronization to music beats. The TIF value remains consistent across models, as it primarily depends on the dancer positions generated by the Dance-Trajectory Navigator.
FP enhances the extraction of group-level information while slightly compromising individual fidelity. This trade-off significantly improves visual performance, as illustrated in Figure \ref{fig:fusion_projection}, which compares EDGE's generation results with and without our module. Although EDGE was initially designed for single-dancer generation and often encounters dancer ambiguity, the integration of FP effectively mitigates this issue, underscoring the module's impact.

\textbf{Effectiveness of Footwork Adaptor.}
\label{sec:faab}
We highlight the importance of natural footwork in enhancing visual aesthetics \cite{edge}. As shown in Table \ref{tab:ablationstudy}, our Footwork Adaptor markedly enhances the model's PFC value. Moreover, when utilized independently, it boosts the model's GMC, underscoring the FA's dual benefits: preventing foot slipping and augmenting dance coherence.

\section{Conclusion}
\noindent This paper introduces TCDiff, a novel framework for high-quality multi-dancer movement generation, along with the concept of dancer ambiguity to guide future research. Our lightweight Fusion Projection module effectively mitigates dancer ambiguity with minimal computational cost. 
Experiments confirm our model's superior performance.

\section{Acknowledgments}
\noindent This work was supported by the National Science Fund of China under Grant Nos. U24A20330, 62361166670 and 62072242.


\small
\bibliography{aaai25}

\end{document}